\documentclass[11pt]{article} 

\usepackage[utf8]{inputenc} 
\usepackage{amsmath}

\usepackage{geometry} 
\usepackage{graphicx} 

\usepackage{booktabs} 
\usepackage{array} 
\usepackage{paralist} 
\usepackage{verbatim} 
\usepackage{subfig} 

\usepackage{fancyhdr} 
\usepackage{sectsty}
\allsectionsfont{\sffamily\mdseries\upshape} 

\usepackage[nottoc,notlof,notlot]{tocbibind} 
\usepackage[titles,subfigure]{tocloft} 

\title{Face  Prediction Model  for an Automatic Age-invariant Face Recognition System}
\author{Poonam Yadav \\ London e-Science Centre\\ Imperial College London\\Email: p.yadav@imperial.ac.uk}
\date{} 

\begin{document}
\maketitle
\begin{abstract}
Automated face recognition and identification softwares are becoming part of our daily life; it finds its abode  not only with Facebook's auto photo tagging, Apple’s iPhoto, Google’s Picasa, Microsoft's Kinect,  but also in Homeland Security Department's dedicated biometric face detection systems. Most of these automatic face identification systems fail where the effects of aging come into the picture.  Little work exists in the literature on the subject of face prediction that accounts for aging, which is a vital part of the computer face recognition systems. In recent years,  individual face components' (e.g. eyes, nose, mouth) features based matching algorithms  have emerged, but these approaches are still not efficient. Therefore, in this work we describe a Face Prediction Model (FPM), which  predicts human face aging or growth related image variation using Principle Component Analysis (PCA)  and  Artificial Neural Network (ANN) learning techniques. The FPM captures the facial changes, which occur with human aging and predicts the facial image with a few years of gap with an acceptable accuracy of face matching from 76 to 86\%.
\end{abstract}
\section{Overview of Face Prediction Model (FPM) } 
In recent years, the requirements of an efficient face detection system triggered a handful of age-invariant face detection algorithms \cite{Lanitis2002, Lanitis2004, Wu2004}.  Sergio Verdict et. al \cite{Verdict} suggested the use of Principle Component Analysis and least square extrapolation techniques for face prediction and presented preliminary study without any conclusion; however, this paper set up a path for research in this direction. The challenge of developing how to model face variations under realistic conditions such as natural aging still remains unsolved. Individual face components' (e.g. eyes, nose, mouth) features based matching algorithms have emerged, but these approaches are still not efficient \cite {Lanitis2008, Otto2012, Bonnen2012}.  The proposed FPM tries to capture the inherent facial features using supervised learning algorithms. The FPM requires the time series image data to capture the temporal and spatial correlation between the images to predict the next time series  image, for example, we provide FPM 4 facial images with 3 years intervals (when subject was 15, 18, 21 and 24 years old), it uses these facial images to train itself for the specific subject and then predicts a facial image of the same subject.  The predicted facial image is matched with the actual facial image at 27 year of the subject; the  matched images' differences are provided as  feedback into the model again, this whole process is part of the ANN back-propagation learning algorithm.

The face images are provided as input  to the FPM. The FPM treats "pixel values" of the face images as "variables" and produces global features in the sense that predicted image is influenced by every input image. PCA is applied to Short Time Fourier Transform (STFT) of  normalised images to find the global features, which generates Global Feature Vectors (GFVs). The ANN architecture, which uses the Back-Propagation (BP) learning algorithm, is used to capture the variations in the input images by feeding the GFVs calculated by PCA and it predicts the face image as an output.
\begin{figure}[!htb]
\includegraphics[width=\linewidth,height = 30mm]{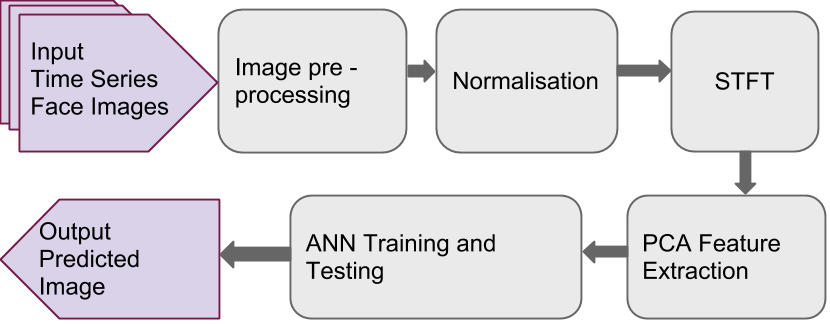}
\caption{Sequential Processing Steps of  the Face Prediction Model}\label{fig:p1}
\end{figure}
The major steps of FPM are shown in Figure 1.
\begin{figure}[ht]
\mbox{
\centerline{\includegraphics[width=\linewidth, height = 20mm]{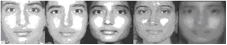}}
}
\caption{The first 4 images from left  are input images with time intervals 2-3 years, and 5th (right most) is the output image by  the Face Prediction Model.}
\label{fig:P5}
\end{figure}

\section{Preliminary Results and Conclusions}
The time series  image data for testing and validation of the FDM is collected by the author of a few known persons and also from fgnet database \cite{fgnet} and FPM Prototype is implemented in Matlab \footnote{FPM matlab source code : https://github.com/pooyadav/FacePredict }. The predicted output image  (last image) in Figure 2 shows the FPM performance. The accuracy of the match varied for different subjects from 78 to 86\% when images are taken with  2-3 years time intervals. The results could be improved by improving the image preprocessing techniques using wavelet transform as it captures localized features better. The Independent component analysis (ICA) will capture more variations in the images than the current approach. Further work is needed to establish appropriate ANN model for the input images separated by larger time gaps ($ > 5$ years).
\newpage
\bibliographystyle{}
\bibliography{}

\begin{thebibliography} {widest entry}
\bibitem[1] {fgnet}http://www.cvmt.dk/projects/fgnet/.
\bibitem [2]{Lanitis2002}A. Lanitis, C. J. Taylor and T. F. Cootes, “Toward Automatic Simulation of Aging Effects on Face Images,” IEEE Transactions on Pattern Analysis and Machine Intelligence, Vol. 24, No. 4, pp 442-455, April 2002.
\bibitem[3]{Lanitis2004}A. Lanitis, C Dragonova and C. Christondoulou, “Comparing Different Classifiers for Automatic Age Estimation,” IEEE Transactions on Systems, Man and Cybernetics, Vol. 34, No.1, pp 621-628, February 2004.
\bibitem[4]{Wu2004}Y. Wu, N. M. Thalmann and D. Thalmann, “A dynamic Wrinkle Model in Facial Animation and Skin aging,” www.miralab.unige.ch/papers/112.pdf, 04/23/04.
\bibitem[5]{Verdict} Sergio Verdict and Hanqi Zhuang,“A preliminary study on human face prediction”, Technical Report, University  of Florida Atlantic University, Boca Raton, FL 33433, 
\bibitem[6]{Starovoitov2002}V.V. Starovoitov, D.I Samal, D.V. Briliuk, “Three Approaches For Face Recognition,” The International Conference on Pattern Recognition and Image Analysis, pp 707-711, October 21-26, 2002.
\bibitem[7]{Lanitis2008} Lanitis, A., "Evaluating the performance of face-aging algorithms," IEEE International Conference on Automatic Face and Gesture Recognition,  pp 1-6,  Sept 17-18, 2008.
\bibitem[8]{Otto2012} Charles Otto, Hu Han and  Anil Jain,"How does aging affect facial components?," The 12th international conference on Computer Vision - Volume 2, pp 189-198, 2012.
\bibitem[9]{Bonnen2012}Kathryn Bonnen and Brendan Klare and Anil K. Jain,"Component-Based Representation in Automated Face Recognition," IEEE Transactions on Information Forensics and Security-Volume 8, pp 239-253, 2013.
\end{thebibliography}

\end{document}